%% file: main.tex
\documentclass[runningheads]{llncs}
\usepackage{graphicx}
\usepackage{amsfonts}
\usepackage{caption}
\usepackage{subcaption}
\usepackage{footnote}
\usepackage{enumerate}
\usepackage{todonotes}
\usepackage{xcolor}
\usepackage{footnote}
\usepackage{xparse}
\usepackage{cite}
\usepackage{booktabs}
\usepackage{url}

\usepackage[hidelinks]{hyperref}
\hypersetup{breaklinks=true,
    colorlinks=true,       % false: boxed links; true: colored links
    linkcolor=red,          % color of internal links (change box color with linkbordercolor)
    citecolor=purple,        % color of links to bibliography
    filecolor=purple,         % color of file links
    urlcolor=magenta        % color of external links
}
\usepackage{amssymb}
\usepackage{amsmath}
\usepackage{keyval}
\usepackage{xspace}
\usepackage{paralist}
\usepackage{listings}
\usepackage{multirow}
\usepackage{stmaryrd}
\usepackage{adjustbox} % scaling figure size
\usepackage{makecell}
\usepackage{sidecap}
\usepackage{booktabs} % For formal tables
\usepackage{threeparttable, tablefootnote}
\usepackage{multirow}
\usepackage{xcolor}
\usepackage{comment}

\RequirePackage{tikz}
\usetikzlibrary{arrows,automata,shapes,calc,through,decorations.pathmorphing,decorations.fractals,chains,shapes.multipart}

\usepackage{arydshln} % hdashline
\usepackage[free-standing-units]{siunitx}
\usepackage{circuitikz}
\usepackage{tabularx} %An extended version of tabular.
\newboolean{blueMode}
\setboolean{blueMode}{false}
\ifthenelse{\boolean{blueMode}}{
  \newcommand*{\newblue}{\textcolor{blue}}
}
{
  \newcommand{\newblue}{}
}
\begin{document}
\title{Robustness Verification of Deep Neural Networks using Star-Based Reachability Analysis with Variable-Length Time Series Input}
\titlerunning{Star-Based Reachability Analysis of DNNs with Time series Data}
\author{Neelanjana Pal\inst{1} \and
% \orcidID{0000-0002-5978-8168} \and
Diego Manzanas Lopez\inst{1} \and
%\orcidID{0000-0002-5978-8168} \and
Taylor T Johnson\inst{1}
%\orcidID{1111-2222-3333-4444}
} 
\authorrunning{Pal et al.}
\institute{Institute for Software Integrated Systems, Vanderbilt University,  Nashville, TN 37212, USA
% \email{neelanjana.pal@vanderbilt.edu}\\
\email{\{neelanjana.pal,diego.manzanas.lopez,taylor.johnson\}@vanderbilt.edu}}
\maketitle              % typeset the header of the contribution
\begin{abstract}
Data-driven, neural network (NN) based anomaly detection and predictive maintenance are emerging research areas. NN-based analytics of time-series data offer valuable insights into past behaviors and estimates of critical parameters like remaining useful life (RUL) of equipment and state-of-charge (SOC) of batteries. However, input time series data can be exposed to intentional or unintentional noise when passing through sensors, necessitating robust validation and verification of these NNs. This paper presents \newblue{a case study} of the robustness verification approach for time series regression NNs (TSRegNN) using set-based formal methods. It focuses on utilizing variable-length input data to streamline input manipulation and enhance network architecture generalizability. The method is applied to two data sets in the Prognostics and Health Management (PHM) application areas: (1) SOC estimation of a Lithium-ion battery and (2) RUL estimation of a turbine engine. The NNs' robustness is checked using star-based reachability analysis, and several performance measures evaluate the effect of bounded perturbations in the input on network outputs, i.e., future outcomes. Overall, the paper offers \newblue{a comprehensive case study} for validating and verifying NN-based analytics of time-series data in real-world applications, emphasizing the importance of robustness testing for accurate and reliable predictions, especially considering the impact of noise on future outcomes.

\keywords{Predictive Maintenance \and
Time Series Data \and
Neural Network Verification \and
Star-set \and
Reachability Analysis \and
Noise \and
Robustness Verification \and
Prognostics and Health Management \and
}
\end{abstract}

\input{1_Introduction}
\input{3_preliminaries}
\input{4_ProblemFormulation}

\input{5_ExperimentalSetup}
\input{6_Evaluation}
\input{7_conclusion}

\bibliographystyle{splncs04}
\bibliography{0_references}

\appendix
\input{10_appendix}

\end{document}

%% file: 1_Introduction.tex
\section{Introduction}
\label{sec: Introduction}

% \todo{\diego{FORMATS has a limit of 15 pages (not including references), keep that in mind}}

Over time, Deep Neural Networks (DNNs) have shown tremendous potential in solving complex tasks, such as image  classification, face detection, object detection, speech recognition, natural language processing, document analysis, etc., sometimes even outperforming humans\cite{lawrence1997face,krizhevsky2012imagenet,lecun1998gradient}. This has motivated a spurt in investigating the applicability of DNNs in numerous real-world  applications, such as biometrics authentication, face authentication for mobile locking systems, malware detection, different bioinformatics applications, etc. In dealing with such susceptible information in these critical areas, safety, security, and verification thereof have become essential design considerations.

Unfortunately, it has been demonstrated that state-of-the-art well-trained networks can be easily deceived by minimal perturbations in the input leading to erroneous predictions \cite{moosavi2016deepfool,LBFGS,goodfellow2014explaining}. The most researched domain for verification of such networks involves image inputs, particularly safety and robustness checking of various classification neural networks \cite{tran2021robustness,anderson2019optimization,botoeva2020efficient,katz2019marabou,mohapatra2020towards,tran2020verification}. Previous research has analyzed feed-forward neural networks (FFNN\cite{tran2019star}), convolutional neural networks (CNN\cite{tran2020verification}), and semantic segmentation networks (SSN\cite{tran2021robustness}) using different set-based reachability tools, such as Neural Network Verification (NNV\cite{tran2020nnv, lopez2023nnv}) and JuliaReach \cite{bogomolov2019juliareach}, among others. 

Input perturbations are not only confined to image-based networks but also have been extended to other input types, including time series data or input signals with different noises in predictive maintenance applications \cite{delillo1999white,truax1999handbook}. One such use case is in the manufacturing industry, where data from process systems, such as IoT sensors and industrial machines, are stored for future analysis \cite{semenick2000time,ferguson1965time}. Data analytics in this context provide insights and statistical information and can be used to diagnose past behavior \cite{zhang2021fault,lv2017weighted}, and predicts future behavior \cite{susto2016dealing,borgi2017data,lin2019time}, maximizing industry production. This application is not only limited to manufacturing, but is also relevant in fields like healthcare digitalization \cite{zeger2006time,touloumi2004analysis} and smart cities \cite{stubinger2020understanding,soomro2019smart}. Noisy input data, here, refers to data containing errors, uncertainties, or disturbances, caused by factors like sensor measurement errors, environmental variations, or other noise sources.
% A time series is a collection of observations of a specific process over time, where the time steps are generally equally spaced (time unit being: minutes, hours, days, weeks, months, etc. and in some cases small deviations in time intervals are acceptable).

While NN applications with image data have received significant attention, little work has been done in the domain of regression-type model verification, particularly with time series data in predictive maintenance applications. Regression-based models with noisy data are crucial for learning data representations and predicting future values, enabling fault prediction and anomaly detection in high-confidence, safety-critical systems \cite{de2022anomaly, kauffman2021palisade}. This motivated us to use verification techniques to validate the output of regression networks and ensure that the output(s) fall within a specific safe and acceptable range.

\paragraph*{Contributions.}
\begin{enumerate}
    \item \newblue{In this paper, we primarily focus on exploring a new case study, specifically examining time-series-based neural networks in two distinct industrial predictive maintenance application domains. We utilize the established concept of star-set-based reachability methods to analyze whether the upper and lower bounds of the output set adhere to industrial guidelines' permissible bounds.} We develop our work\footnote{The code is available at: \url{https://github.com/Neelanjana314/nnv/tree/master/code/nnv/examples/Submission/FMCIS2023}} as an extension of the NNV tool to formally analyze and explore regression-based NN verification for time series data using sound and deterministic reachability methods and experiment on different discrete time signals to check if the output lies within pre-defined safe bounds.
    \item \newblue{Another significant contribution of our work is the flexibility of variable-length inputs in neural networks. This approach simplifies input manipulation and enhances the generalizability of network architectures. Unlike published literature that relied on fixed-sized windows \cite{ForMuLA, muller2022third}, which necessitated preprocessing and experimenting with window sizes, our method allows for flexibility in utilizing any sequence length. This flexibility improves the generalizability of reachability analysis.}
    \item We run an extensive evaluation on two different network architectures in two different predictive maintenance use cases. \newblue{In terms of evaluation, we have introduced a novel robustness measure called Percentage Overlap Robustness (POR). Unlike the existing Percentage Sample Robustness (PR/PSR) \cite{tran2021robustness}, which considers only instances where reachable bounds remain entirely within permissible bounds, the proposed POR accounts for all instances with overlap.}
    \item Finally, we develop insights on evaluating the reachability analysis on those networks and possible future direction.
\end{enumerate}

\paragraph*{Outline.} 
The paper is organized as follows:
Section \ref{sec: Preliminaries} provides the necessary context for the background; Section \ref{Sec: Advnoise} details the adversarial noises; Section \ref{Sec: VerProp} defines the verification properties; Section \ref{Sec:ReachabilityLayers} explains the reachability calculations for layers to accommodate variable-length input; Section \ref{sec: problem} defines the research problem, and Section \ref{sec: Experimental Setup} describes the methodology, including dataset, network models, and input attacks. Section \ref{sec: Evaluation} presents the experimental results, evaluation metrics, and their implications. Finally, Section \ref{sec: Conclusion and Future Work} summarizes the main findings and suggests future research directions.

%% file: 3_preliminaries.tex
\section{Preliminaries}
\label{sec: Preliminaries}
\noindent{This section introduces some basic definitions and descriptions necessary to understand the progression of this paper and the necessary evaluations on time series data.}
% \todo{add an overview sentence of what this section is about}
 
\subsection{Neural Network Verification Tool and Star Sets}
The Neural Network Verification (NNV) tool is a framework for verifying the safety and robustness of neural networks \cite{tran2020nnv, lopez2023nnv}. It analyzes neural network behavior under various input conditions, ensuring safe and correct operation in all cases. NNV supports reachability algorithms like the over-approximate star set approach \cite{tran2020verification,tran2019fm}, calculating reachable sets for each network layer. These sets represent all possible network states for a given input, enabling the verification of specific safety properties. NNV is particularly valuable for safety-critical applications, such as autonomous vehicles and medical devices, ensuring neural networks are trustworthy and reliable under all conditions, maintaining public confidence. 
For this paper, we have implemented the work as an extension of NNV tool and used the star [Def. \ref{def:star}] based reachability analysis to get the reachable sets of the neural networks at the outputs.
\newblue{\begin{definition}\label{def:star} \emph{\textbf{A generalized star set}} (or simply star) $\Theta$ is a tuple $\langle c, V, P \rangle$ where $c \in \mathbb{R}^n$ is the center, $V = \{v_1, v_2, \cdots, v_m\}$ is a set of m vectors in $\mathbb{R}^n$ called basis vectors, and $P: \mathbb{R}^m \to \{ \top, \bot\}$ is a predicate. The basis vectors are arranged to form the star's $n \times m$ basis matrix. The set of states represented by the star is given as:
\begin{equation}
 \llbracket \Theta \rrbracket = \{x~|~x = c + \Sigma_{i=1}^m(\alpha_iv_i)~\text{and}~P(\alpha_1, \cdots, \alpha_m) = \top \}.
\end{equation}
In this work, we restrict the predicates to be a conjunction of linear constraints, $P(\alpha) \triangleq C\alpha \leq d$ where, for $p$ linear constraints, $C \in \mathbb{R}^{p \times m}$, $\alpha$ is the vector of $m$-variables, i.e., $\alpha = [\alpha_1, \cdots, \alpha_m]^T$, and $d \in \mathbb{R}^{p \times 1}$.% A star is an empty set, i.e., $\Theta = \emptyset$ if and only if the predicate $P(\alpha)$ is infeasible. In other words, we can say the predicate polyhedron $P(\alpha)$ is an empty set, i.e., $P(\alpha) = \emptyset$.
\end{definition}
\vspace*{-\baselineskip}
\vspace*{-\baselineskip}
\begin{figure}[h!]
    \centering
    \includegraphics[width = 0.8\columnwidth]{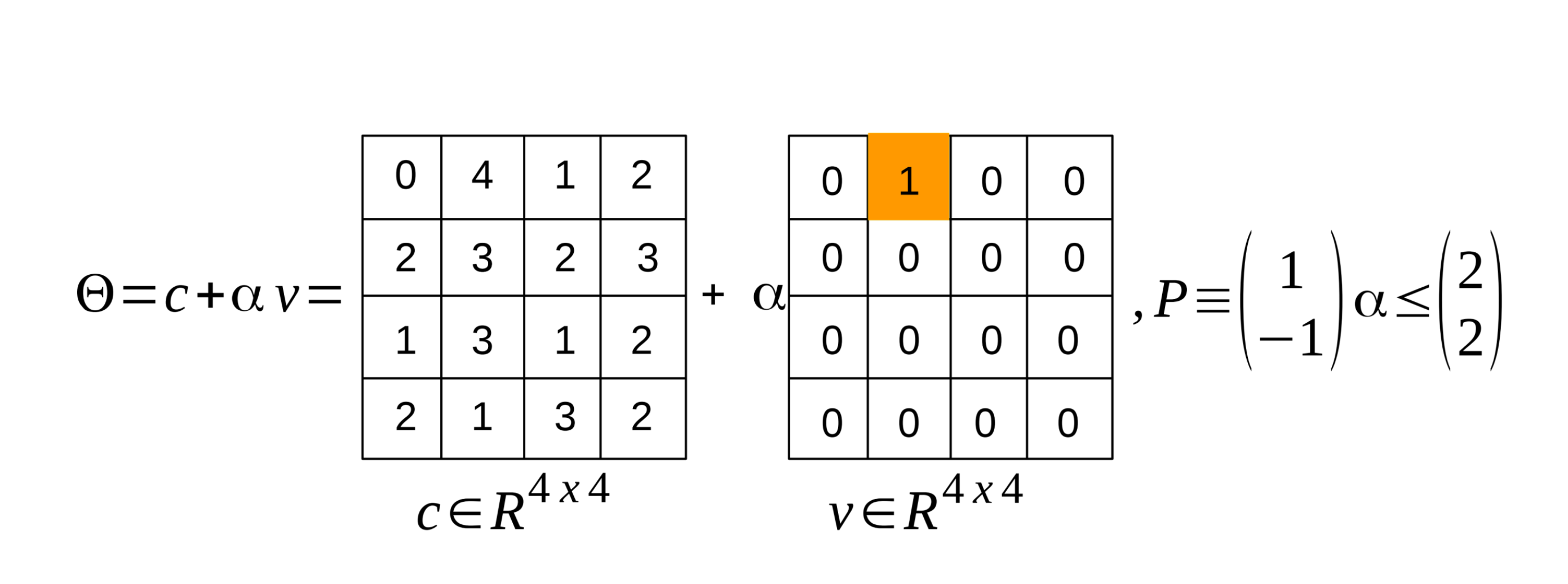}
    \caption{Star for Time-series Data with four Feature Values (rows) with four time-steps (columns)}
    \label{fig:Signalstar}
\end{figure}
 \vspace*{-\baselineskip}
% \end{definition}
An alternative approach to defining a Star set for time series data involves using the upper and lower bounds of the noisy input while centering the actual input. These bounds on each input parameter, along with the predicates, create the complete set of constraints the optimizer will solve to generate the initial set of states.}

A 4 × 4 time series data with a bounded disturbance
$b \in [-2, 2]$ applied on the time instance 2 of feature 1, i.e., position (1, 2) can be described as a Star depicted in Fig. \ref{fig:Signalstar}.
\subsection{Time Series and Regression Neural Network}
\textbf{Signal.} The definition of a `signal' varies depending on the applicable fields. In the area of signal processing, a \emph{\textbf{signal}} $S$ can be defined as some physical quantity that varies with respect to (w.r.t.) some independent dimension (e.g., space or time) \cite{priemer1991introductory}. 
In other words, a signal can also be thought of as a function that carries information about the behavior of a system or properties of some physical process \cite{priemer1991signals}. 
\begin{equation}
    \label{equ: Signal}
     S = g(q) 
\end{equation}
where $q$ is space, time, etc.
Depending on the nature of the spaces signals are defined over, they can be categorized as discrete or continuous. Discrete time signals are also known as time series data.
% \end{definition}

\noindent{We next define the specific class of signals considered in this paper, namely time series.}

\begin{definition} \label{def:Timeseries}
A \textbf{time series signal} $S_T$ is defined as an ordered sequence of values of a variable (or variables) at different time steps. In other words, a time series signal is an ordered sequence of discrete-time data of one or multiple features\footnote{Each \textbf{feature} is a measurable piece of data that is used for analysis.}.
\begin{equation}\label{equ: Timeseries}
\begin{split}
    S_T = s_{t_1}, s_{t_2}, s_{t_3}, ... \\
    T = t_1, t_2, t_3, ... 
\end{split}
\end{equation}
where, $t_1,t_2,t_3,\hdots$ is an ordered sequence of instances in time $T$ and $S_T = s_{t_1},s_{t_2},s_{t_3},\hdots$ are the signal values at those time instances for each $t = t_i$.

Here, sometimes we have used `signal' to refer to the `time series signal.'
\end{definition}
\noindent{Next, we define the specific types of neural networks considered in this paper, namely regression neural networks (specifically time series regression neural networks).}

\begin{definition} A \textbf{time series regression neural network (TSRegNN)} $f$ is a nonlinear/partially-linear function that maps each time-stamped value $x(i, j)$ (for $i^{th}$ feature and $j^{th}$ timestamp) of a single or multifeatured time series input $\textbf{x}$ to the output $\textbf{y}$.
% = \{y_1, y_2,\ldots,y_p\}$, $p$ is the number of expected values (either single or multiple (application dependent)) at the output.
\begin{equation}
    \label{equ: TSNN}
    f:~\textbf{x} \in \mathbb{R}^{\newblue{n_f}\times t_s} \rightarrow \textbf{y} \in {\mathbb{R}}^{p \times q} \\
\end{equation}
where $t_s, \newblue{n_f}$ are the time-sequence length and the number of features of the input data, respectively, $(j, i) \in \{ 1,\ldots,t_s \} \times \{ 1,\ldots,\newblue{n_f} \}$ are the time steps and corresponding feature indices, respectively, and $p$ is the number of values present in the output, while $q$ is the length of each of the output values; \newblue{it can either be equal to $t_s$ or not, depending on the network design.} 

Here, each row of $\textbf{x}$ represents a timestamped feature variable.
\end{definition}

\subsection{Reachability of a Time Series Regression Network}\label{Sec:Reachability}
In this section, we provide a description of how the reachability of a NN layer and the NN as a whole is computed for this study.
\begin{definition}\label{Def: Layer} A \textbf{layer} $L$ of a TSRegNN is a function $h:~u \in \mathbb{R}^{j} \rightarrow v \in {\mathbb{R}}^{p},$ with input $u \in {R}^{j}$ and output $v \in {R}^{p}$ defined as
follows 
\begin{equation}
    \label{equ: NNlayer}
    v = h(u)
\end{equation}
where the function $h$ is determined by parameters $\theta$, typically defined as a tuple $\theta = \langle \sigma, W, b \rangle$ for fully-connected
layers, where $ W \in {R}^{j\times p}, b \in {R}^{p}$, and activation function $\sigma :
{R}^{j} \rightarrow {R}^{p}$. Thus, the fully connected NN layer is described
as
\begin{equation}
    \label{equ: FC}
    v = h(u) = \sigma(\textbf{W}\times u + \textbf{b})
\end{equation}

For convolutional NN-Layers, $\theta$ may include parameters like the filter size, padding, or dilation factor, and the function in Eq. \ref{equ: FC} may need alterations.

\end{definition}

\begin{definition} Let $h:~u \in \mathbb{R}^{j} \rightarrow v \in {\mathbb{R}}^{p}$, be a NN layer as described in Eq. \ref{equ: NNlayer}. The
\textbf{reachable set} ${R_h}$, with input, $I \in {R}^{n}$ is defined as
\begin{equation}
    \label{equ: ReachLayer}
    \mathcal{R}_{h} \triangleq \{v~|~v = h(u),~u \in \mathcal{I}\}
    \end{equation}
\end{definition}

\noindent \textbf{Reachability analysis (or shortly, reach) of a TSRegNN} $f$ on Star input set $I$ is similar to the reachable set calculations for CNN\cite{tran2020verification} or FFNN\cite{tran2019star}, the only difference being both the previous works had been done for classification networks. 
\begin{equation}
\begin{split}
	Reach(f, I):~&I \rightarrow \mathcal{R}_{ts} %\\
				 %&x \rightarrow y = f(x). 
\end{split}
\end{equation}
We call $\mathcal{R}_{ts}(I)$ the \emph{output reachable set} of the TSRegNN corresponding to the input set $I$.

For a regression type NN, the output reachable set can be calculated as a step-by-step process of constructing the reachable sets for each network layer. 
 \begin{equation*}
    \begin{split}
      \mathcal{R}_{L_{1}} &\triangleq \{v_1~|~v_1 = h_1(x),~ x \in \mathcal{I}\}, \\
      \mathcal{R}_{L_{2}} &\triangleq \{v_2~|~v_2 = h_2(v_1),~ v_1 \in \mathcal{R}_{L_1}\}, \\
     &\vdots \\
    \mathcal{R}_{ts} = \mathcal{R}_{L_k} &\triangleq \{v_k ~|~ v_k = h_k(v_{k-1}),~v_{k-1} \in \mathcal{R}_{L_{k-1}}\}, \\
    \end{split}
  \end{equation*}
  where $h_k$ is the function represented by the $k^{th}$ layer $L_k$. The reachable set $\mathcal{R}_{L_k}$ contains all outputs of the neural network corresponding to all input vectors $x$ in the input set $\mathcal{I}$.
\section{Adversarial Noise}\label{Sec: Advnoise} In the case of time series samples, while the sensor transmits the sampled data, sensor noises might get added to the original data. One example of such noise is sensor vibration, but sometimes the actual sources are not even known by the sensor providers\cite{martinez2018ihorizon}.
 
\begin{definition}
A \textbf{noise} can be defined as some unintentional, usually  small-scaled signal which, when added to the primary signal, can cause malfunctioning of the equipment in an industrial premise. Mathematically, a noisy signal $s^{noise}$ 
%$=$ $[s_1^{noise}$, $\dots$, $s_n^{noise}]$ 
can be produced by a linear parameterized function $g_{\epsilon,s^{noise}}(\cdot)$ that takes an input signal and produces the corresponding noisy signal. 
\begin{equation}
	s^{noise} = g_{\epsilon, s^{noise}}(s) = s + \Sigma_{i=1}^n\epsilon_i \cdot s_i^{noise}
\end{equation} 
% where $s_i^{noise}$ is the 
For time series data, we can also assume the noise as a set of unit vectors associated with a coefficient vector $\epsilon$ at each time step $i$, where the value of the coefficient vector $\epsilon$ is unknown but bounded within a range $[\underline{\epsilon}, \overline{\epsilon}]$, i.e., $\underline{\epsilon_i} \leq \epsilon_i \leq \overline{\epsilon_i}$.
\end{definition}
% \diego{Should we add to the definition something that shows that the noise/perturbation is some value $\epsilon$ $\leq$ $\epsilon_{max}$ such that the value of the data $x_i$ in set $X$, |$x_i$ + $\epsilon$| $\leq$ |$x_i$ + $\epsilon_{max}$|}
\vspace*{-\baselineskip}
\subsubsection{Types of Possible Noises.}\label{Noise Types} For an input sequence with $t_s$ number of time instances and $n_f$ number of features, there can be four types of noises ($l_\infty$ norm) [\ref{l_inf norm}], based on its spread on the signal. They can be categorized as below:
\begin{enumerate}

    \item \textbf{Single Feature Single-instance Noise (SFSI)} i.e., perturbing a feature value only at a particular instance ($t$) by a certain percentage around the actual value.
    \begin{equation}\label{equ: SFSI}
	s^{noise} = g_{\epsilon, s^{noise}}(s) = s + \epsilon_t \cdot s_t^{noise}
\end{equation}
    \item \textbf{Single Feature All-instances Noise (SFAI)} i.e., perturbing a specific feature throughout all the time instances by a certain percentage around the actual values of a particular feature.
        \begin{equation}\label{equ: SFAI}
	s^{noise} = g_{\epsilon, s^{noise}}(s) = s + \Sigma_{i=1}^n\epsilon_i \cdot s_i^{noise}
\end{equation}
    \item \textbf{Multifeature Single-instance Noise (MFSI)} i.e., perturbing all feature values but only at a particular instance (t), following Eq.~\ref{equ: SFSI} for all features.

    \item \textbf{Multifeature All-instance Noise (MFAI)} i.e., perturbing all feature values throughout all the instances, following Eq.~\ref{equ: SFAI} for all features.  
 \end{enumerate}

A sample plot for all four types of noises is shown in [\ref{AppendixNoise}].

\section{Verification Properties}\label{Sec: VerProp} Verification properties can be categorized into two types: local properties and global properties. A local property is defined for a specific input \newblue{$x$ at time-instance $t$ or a set of points $X$} in the input space \textbf{\newblue{$R^{n_f \times t_s}$}}. In other words, a local property must hold for certain specific inputs. On the other hand, a global property \cite{wang2022tool} is defined over the entire input space \newblue{$R^{n_f \times t_s}$} of the network model and must hold for all inputs without any exceptions.
\subsubsection{Robustness.}
Robustness refers to the ability of a system or a model to maintain its performance and functionality under various challenging conditions, uncertainties, or perturbations. It is a desirable quality that ensures the system's reliability, resilience, and adaptability in the face of changing or adverse circumstances. For an input perturbation measured by $\delta$ and admissible output deviation $\epsilon$, the `delta-epsilon' formulation for the desired robustness property can be written as:
\begin{equation}
    ||x' - x||_{\infty} < \delta \implies || f(x') - f(x) ||_{\infty} < \epsilon
\end{equation}
where $x$ is the original input belonging to the input space \newblue{$R^{n_f \times t_s}$}, $x'$ is the noisy input, $f(x')$ and $f(x)$ are NN model outputs for, respectively, $x'$ and $x$, $\delta$ is the max measure of the noise added, $\epsilon$ is the max deviation in the output because of the presence of noise ($\delta, \epsilon \in \textbf{R} >0 )$.
\paragraph*{\textbf{Local Robustness.}}\label{Robust} Given a TSRegNN $f$ and an input time series signal $S$, the network is called \textbf{locally robust} to any noise $\mathcal{A}$ if and only if: the estimated output reachable bounds for a particular time-step corresponding to the noisy input lie between predefined allowable bounds w.r.t to the actual signal.

\textbf{Robustness Value (RV)} of a time series signal $S$ is a binary variable, which indicates the local robustness of the system. RV is $1$ when the estimated output range for a particular time instance (t) lies within the allowable range, making it locally robust at t; otherwise, RV is 0. 

$RV = 1 \iff {LB_t}^{est} \geq {LB_t}^{allow} \newblue{\land} {UB_t}^{est} \leq {UB_t}^{allow}$
else, RV = 0 

where ${LB_t}^{est}$ and ${UB_t}^{est}$  are estimated bounds and ${LB_t}^{allow}$ and ${UB_t}^{allow}$ are allowable bounds.

\begin{definition} \textbf{Percentage Sample Robustness (PR)}\label{def:PSR} of a TSRegNN corresponding to any noisy input is defined as
\begin{equation}
PR = \frac{N_{robust}}{N_{total}}\times 100\%,
\end{equation}
where $N_{robust}$ is the total number of robust time instances, and $N_{total}$ = the total number of time steps in the time series signal. Percentage robustness can be used as a measure of \textbf{global robustness \newblue{\cite{wang2022tool}}} of a TSRegNN w.r.t any noise.

\newblue{In this study, we adapt the concept of Percentage Robustness (PR) previously used in image-based classification or segmentation neural networks \cite{tran2021robustness} to time-series inputs. PR in those cases assessed the network's ability to correctly classify/segment inputs even with input perturbations for a given number of images/pixels. We extend this concept to analyze the robustness of time-series inputs in our research.}

\end{definition}

\begin{definition} \textbf{Percentage Overlap Robustness (POR)}\label{def:POR} of a TSRegNN corresponding to any noisy input is defined as
\begin{equation}
POR = \frac{\Sigma_{i=1}^{N_{total}}(PO_i)}{N_{total}}\times 100\%,
\end{equation}
where $N_{total}$ = total number of time instances in the time series signal, and $PO_i$ is the percentage overlap between estimated and allowed ranges at each time step w.r.t the estimated range
\begin{equation}
 PO = \frac{Overlapped~Range}{Estimated~Range}
\end{equation}
Here $Overlapped~Range$ is the overlap between the estimated range and the allowable range for a particular time step, whereas $Estimated~Range$ is the output estimation given by the TSRegNN for that time step. 
Percentage overlap robustness can also be used as a measure of \textbf{global robustness \newblue{\cite{wang2022tool}}} of TSRegNN.
\end{definition}
\newblue{When selecting robustness properties, it is crucial to consider the specific application area. If the application allows for some flexibility in terms of performance, POR can be utilized. On the other hand, if the application requires a more conservative approach, PR should be considered.}
An example showing calculations for the robustness measures is shown in [\ref{Robustness Measures Example}].
\vspace*{-\baselineskip}
\subsubsection{Monotonicity.} In PHM applications, the monotonicity property refers to the system's health indicator, i.e., the degradation parameter exhibiting a consistent increase or decrease as the system approaches failure. PHM involves monitoring a system's health condition and predicting its Remaining Useful Life (RUL) to enable informed maintenance decisions and prevent unforeseen failures. For detailed mathematical modeling of the monotonicity property, please refer to \cite{sivaraman2020counterexample} and the latest report on formal methods at \cite{ForMuLA}. In general, for a TSRegNN $f:~\textbf{x} \in \mathbb{R} \rightarrow \textbf{y} \in {\mathbb{R}}$ with single-featured input and output spaces, at any time instance $t$, the property for monotonically decreasing output can be written as:
 \begin{equation}\label{equ: monotonicity}
     \begin{split}
         \forall x' \exists \delta : x \leq x' \leq x+\delta \implies f(x') \leq f(x) \\
         \forall x' \exists \delta : x-\delta \leq x' \leq x \implies f(x') \geq f(x)\\
     \end{split}
 \end{equation}
 This is a local monotonicity property. If this holds true for the entire time range, then the property can be considered as a global property \cite{wang2022tool}. In this paper, the monotonicity property is only valid for the PHM examples for RUL estimation. 
\section{Reachability of Specific Layers to Allow Variable-Length Time Series Input}\label{Sec:ReachabilityLayers}

\subsubsection{Reachability Of A Fully-connected Layer.}
We consider a fully-connected layer with the following parameters: the weights $W_{fc} \in {R}^{op \times ip}$ and the bias $b_{fc} \in {R}^{op \times 1}$, where $op$ and $ip$ are, respectively, the output and input sizes of the layer. The output of this fully connected layer w.r.t an input $i \in {R}^{ip \times T_s}$ will be
\begin{equation*}
    \begin{split}
        o = W_{fc} \times i + b_{fc} \\
where~output~o \in {R}^{op \times T_s}
    \end{split}    
\end{equation*}
Thus, we can see that the layer functionality does not alter the output size for a variable length of time sequence, making the functionality of this layer independent of the time series length.

\noindent The reachability of  a fully-connected layer will be given by the following lemma.
\vspace*{-\baselineskip}
\begin{lemma} \label{lem:fc}
The reachable set of a fully-connected layer with 
 a Star input set $I = \langle c, V, P \rangle$ is another Star $I' = \langle c', V', P' \rangle$ where $c' = W_{fc} \times c + b_{fc}$, the matrix multiplication of $c$ with Weight matrix $W_{fc}$,$V' = \{v'_1,...,v'_m\}$, where $v'_i = W_{fc} \times v_i$, the matrix multiplication of the weight matrix and the $i^{th}$ basis vector, \newblue{and $P' = P$}. 
\end{lemma}
\vspace*{-\baselineskip}
\subsubsection{Reachability of a 1D Convolutional Layer.} We consider a 1d convolution layer with the following parameters: the weights $W_{conv1d} \in {R}^{w_f \times nc \times fl}$ and the bias $b_{conv1d} \in {R}^{1 \times fl}$, the padding size $P$, the stride $S$, and the dilation factor $D$; where $w_f, nc~and~fl$ are the filter size, number of channels and number of filters respectively.

The output of this 1d convolution layer w.r.t an input $i \in {R}^{ip \times T_s}$ will be
\begin{equation*}
    \begin{split}
        o = W'_{conv1d} \cdot i' + b_{conv1d} ~~dot~product~along~time~dimesion~for~each~filter\\
where~output~o \in {R}^{fl \times T'_s}
    \end{split}    
\end{equation*}
where $T'_s = T_s + T_d - T_{fl} $ is the new time series length at the output and $T_d, T_{fl}$ are the time lengths contributed by the dilation factor and the 1d convolution function, respectively. $w'_{conv1d}$ is the modified weight matrix after adding dilation, and $i'$ is the modified input after padding. We can see when $T_d$ becomes equal to $T_{fl}$ for any convolution layer, the layer functionality becomes independent of the length of the time series.

\noindent The reachability of  a 1d convolution layer will be given by the following lemma.
\vspace*{-\baselineskip}
\begin{lemma} \label{lem:conv1d}
The reachable set of a 1d convolution layer with 
 a Star input set $I = \langle c, V, P \rangle$ is another Star $I' = \langle c', V', P' \rangle$ where $c' = W_{conv1d} \cdot c $, 1d convolution applied to the \newblue{basis} vector $c$ with Weight matrix $W_{conv1d}$,$V' = \{v'_1,...,v'_m\}$, where $v'_i = W_{conv1d} \cdot v_i$, is the 1d convolution operation with zero bias applied to the generator vectors, i.e., only using the weights of the layer, \newblue{and $P' = P$}.
\end{lemma}

%% file: 4_ProblemFormulation.tex
\section{Robustness Verification Problem Formulation}\label{sec: problem}
We consider the verification of the robustness and the monotonicity properties. 
\begin{problem}[\textbf{Local Robustness Property}]\label{prob:1}
Given a TSRegNN $f$, a time series signal $S$, and a noise $\mathcal{A}$, prove if the network is locally robust or non-robust [Sec. \ref{Robust}] w.r.t the noise $\mathcal{A}$; i.e., if the estimated bounds obtained through the reachability calculations lie within the allowable range of the actual output for the particular time instance.
\end{problem}
\begin{problem}[\textbf{Global Robustness Property}]\label{prob:2}
Given a TSRegNN $f$, a set of $N$ consecutive time-series signal $\textbf{S} = \{S_1, \dots, S_N\}$, and a noise $\mathcal{A}$, compute the percentage robustness values (PR [Def. \ref{def:PSR}] and POR [Def. \ref{def:POR}]) corresponding to $\mathcal{A}$.
\end{problem}
\begin{problem}[\textbf{Local Monotonicity Property}]\label{prob:3}
Given a TSRegNN $f$, a set of $N$ consecutive time-series signal $\textbf{S} = \{S_1, \dots, S_N\}$, and a noise $\mathcal{A}$, show that both the estimated RUL bounds of the network [Eq. \ref{equ: monotonicity}] corresponding to noisy input $S'_t$ at any time instance $t$ are monotonically decreasing.
\end{problem}

To get an idea of the global performance \cite{wang2022tool} of the network, local stability properties have been formulated and verified for each point in the test dataset for 100 consecutive time steps.

 The core step in solving these problems is to solve the local properties of a TSRegNN $f$ w.r.t a noise $\mathcal{A}$. It can be done using over-approximate reachability analysis, computing the `output reachable set' $\mathcal{R}_ts = Reach(f, I)$ that provides an upper and lower bound estimation corresponding to the noisy input set $I$. 

\newblue{In this paper, we propose using percentage values as robustness measures for verifying neural networks (NN). We conduct reachability analysis on the output set to ensure it stays within predefined safe bounds specified by permissible upper-lower bounds. The calculated overlap or sample robustness, expressed as a percentage value, represents the NN's robustness achieved through the verification process under different noise conditions.} The proposed solution takes a sound and incomplete approach to verify the robustness of regression neural networks with time series data. The approach over-approximates the reachable set, ensuring that any input point within the set will always have an output point contained within the reachable output set (sound [Def. \ref{def: soundness}]). However, due to the complexities of neural networks and the over-approximation nature of the approach, certain output points within the reachable output set may not directly correspond to specific input points (incomplete [Def. \ref{def: completeness}]). Over-approximation is commonly used in safety verification and robustness analysis of complex systems due to its computational efficiency and reduced time requirements compared to exact methods.

%% file: 5_ExperimentalSetup.tex
\section{Experimental Setup}
\label{sec: Experimental Setup}
\subsection{Dataset Description}\label{Dataset} For evaluation, we have considered two different time series datasets for PHM of a Li battery and a turbine.

\paragraph{\textbf{Battery State-of-Charge Dataset (BSOC)}\cite{kollmeyer2020lg}:} This dataset is derived from a new 3Ah LG HG2 cell tested in an 8 cu.ft. thermal chamber using a 75amp, 5-volt Digatron Firing Circuits Universal Battery Tester with high accuracy (0.1 of full scale) for voltage and current measurements. The main focus is to determine the State of Charge (SOC) of the battery, measured as a percentage, which indicates the charge level relative to its capacity. SOC for a Li-ion battery depends on various features, including voltage, current, temperature, and average voltage and current. The data is obtained from the `LG\_HG2\_Prepared\_Dataset\_McMasterUniversity\_Jan\_2020', readily available in the dataset folder \cite{kollmeyer2020lg}. The training data consists of a single sequence of experimental data collected while the battery-powered electric vehicle during a driving cycle at an external temperature of 25 degrees Celsius. The test dataset contains experimental data with an external temperature of -10 degrees Celsius.

\paragraph{\textbf{Turbofan Engine Degradation Simulation Data Set (TEDS)}\cite{Prognost61:online,saxena2008turbofan}:} This dataset is widely used for predicting the Remaining Useful Life (RUL) of turbofan jet engines \cite{Prognost61:online}. Engine degradation simulations are conducted using C-MAPSS (Commercial Modular Aero-Propulsion System Simulation) with four different sets, simulating various operational conditions and fault modes. Each engine has 26 different feature values recorded at different time instances. To streamline computation, features with low variability (similar to Principal Component Analysis \cite{pearson1901liii}) are removed to avoid negative impacts on the training process. The remaining 17 features [\ref{AppendixPrognosability},\ref{AppendixSampleDataset}] are then normalized using z-score (mean-standard deviation) for training. The training subset comprises time series data for 100 engines, but for this paper, we focus on data from only one engine (FD001). For evaluation, we randomly selected engine 52 from the test dataset.
 
\subsection{Network Description}

% \paragraph{\textbf{Battery State-of-Charge Dataset (BSOC)}:} 
The network architecture used for training the BSOC dataset, partially adopted from \cite{mathworksPredictBattery}, is a regression CNN, as shown in [Fig~\ref{fig:FigArchitecture},\ref{AppendixNetworks}]. The network has five input features which correspond to one SOC value. Therefore, the TSRegNN for the BSOC dataset can be represented as:
\begin{equation}
    \label{equ: modTSRNN}
    \begin{split}
    f:~x \in \mathbb{R}^{5\times t_s} \rightarrow y \in {\mathbb{R}^{1 \times t_s}} \\
        \hat{SOC}_{t_s} = f(t_s)
    \end{split}
\end{equation}
% \paragraph{\textbf{Turbofan Engine Degradation Simulation Data Set (TEDS)}:} 
The network architecture used for training the TEDS dataset is also a regression CNN, adopted from \cite{mathworksRemainingUseful} and shown in [Fig~\ref{fig:FigArchitecture},\ref{AppendixNetworks}]. The input data is preprocessed to focus on 17 features, corresponding to one RUL value for the engine. Therefore, the TSRegNN for the TEDS dataset can be represented as:
\begin{equation}
    \label{equ: modTSRNN}
    \begin{split}
        f:~x \in \mathbb{R}^{17\times t_s} \rightarrow y \in {\mathbb{R}^{1 \times t_s}} \\
        \hat{RUL}_{t_s+1} = f(t_s)
    \end{split}
\end{equation}
% The $T^{th}$ value in the output represents the desired estimation of RUL,  given the series of past T values of 17 features.
The output's $t_s^{th}$ value represents the desired estimation of SOC or RUL, with the given series of past $t_s$ values for each feature variable.
% \subsection{Verification Properties Considered}
% For all four noise scenarios mentioned in Sec.~\ref{Noise Types}, local and global (here for 100 consecutive time steps) robustness properties are considered for both datasets. An additional local monotonicity property is also considered for the turbine RUL estimation example. 

%% file: 6_Evaluation.tex
\section{Experimental Results and Evaluation}
\label{sec: Evaluation} The actual experimental results shown in this paper are conducted in a Windows-10 computer with the 64-bit operating system, Intel(R) Core(TM) i7-8850H processor, and 16 GB RAM.

For all four noise scenarios [Sec.~\ref{Noise Types}], local and global (for 100 consecutive time steps) robustness properties are considered for both datasets. The local monotonicity property is only considered for the turbine RUL estimation example.

\paragraph{\textbf{Battery State-of-Charge Dataset (BSOC)}:}
In this dataset, the output value (SOC) is supposed to be any value between 0 and 1 (or 0 and 100$\%$). But, for the instances where the lower bound is negative, we instead treat it as 0 because a negative SOC does not provide any meaningful implications. 

For SFSI, for a random (here feature 3) input feature-signal, the noise is added only at the last time step ($t_{30}$) of the 3rd feature, whereas for SFAI, noise is added throughout all the time instances of the input signal. The effect of four different noise values, 1$\%$, 2.5$\%$, 5$\%$ and 10$\%$ of the mean($\mu$), are then evaluated using over-approximate star reachability analysis [Sec.~\ref{Sec:Reachability}] on 100 consecutive input signal, each with 30 time instances. We considered $\pm 5\%$ around the actual SOC value as the allowable bounds. For all the noises, 2 different robustness values, PR [Def.~\ref{def:PSR}] and POR [(Def.~\ref{def:POR}] are then calculated, and comparative tables are shown below in Table~\ref{tab:Table2}. 
% \vspace*{-\baselineskip}
\begin{figure}[t!]
    \centering
    \includegraphics[width = 0.8\columnwidth]{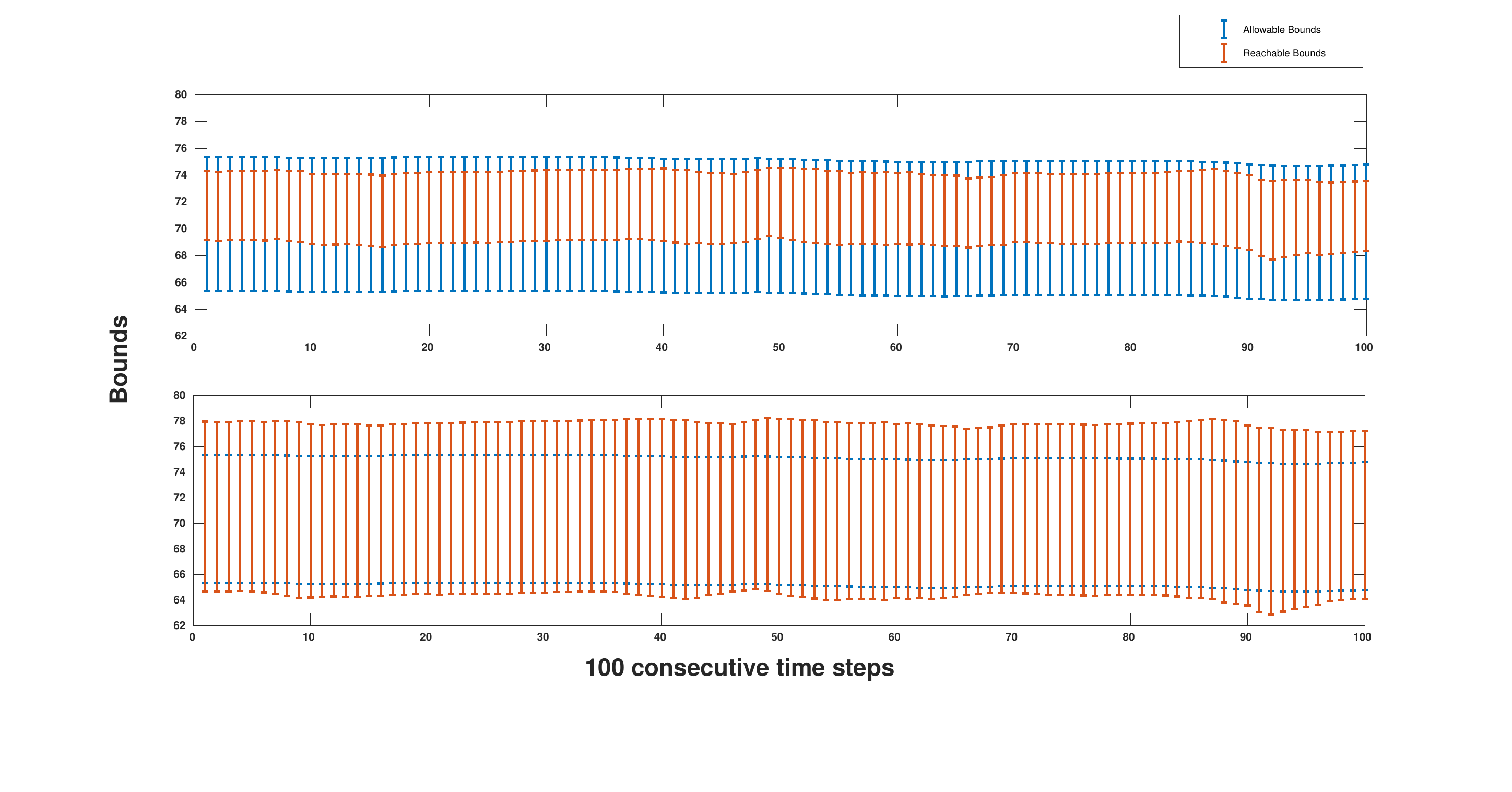}
    \vspace*{-\baselineskip}
    \vspace*{-\baselineskip}
        \caption{Allowable (blue) and reachable (red) bounds for battery SOC dataset for 100 consecutive time steps and 2 different SFAI noise values $1\%$ (upper), and $2.5\%$ (lower) respectively }
    \label{fig:BSOCBounds}
    \vspace*{-\baselineskip}
    % \vspace*{-\baselineskip}
    % \vspace*{-\baselineskip}
\end{figure}
\begin{table}[h!]
    \caption{Global Robustness: Percentage Robustness(PR) for noises for 100 consecutive time steps}
    \label{tab:Table2}
    \centering 
    \begin{tabular}{ccccccc}
    \toprule
    \centering
    $noise$ & $PR_{SFSI}$ & $POR_{SFSI}$ & $avgRT_{SFSI}(s)$ & $PR_{SFAI}$ & $POR_{SFAI}$ & $avgRT_{SFAI}(s)$\\ 
    \hline 
    1 & 100 & 100 & 0.7080 & 100 & 100 & 20.9268\\ 
    2.5 & 100 & 100 & 0.7080 & 100 & 100 & 20.9991\\
    5 & 100 & 100 & 0.7116 & 100 & 100 & 21.0729\\
    10 & 100 & 100 & 0.7027 & 100 & 100 & 21.0780\\
    \hline 
    \end{tabular}
    \begin{tabular}{ccccccc}
    \toprule
    \centering
    $noise$ & $PR_{MFSI}$ & $POR_{MFSI}$ & $avgRT_{MFSI}(s)$ & $PR_{MFAI}$ & $POR_{MFAI}$ & $avgRT_{MFAI}(s)$\\ 
    \hline 
    1 & 100 & 100 & 0.7653 & 100 & 100 & 36.1723\\ 
    2.5 & 0  & 73.87 & 0.8251 & 0  & 73.87 & 59.0588\\
    5 & 0 & 35.95 & 0.9026 & 0 & 35.95 & 91.6481\\
    10 & 0 & 17.89 & 1.1051 & 0 & 17.89 & 163.7568\\
    \hline 
    \end{tabular}
    % \vspace*{-\baselineskip}
\end{table}
\begin{figure}[h!]
    \centering
    \includegraphics[width = 0.9\columnwidth]{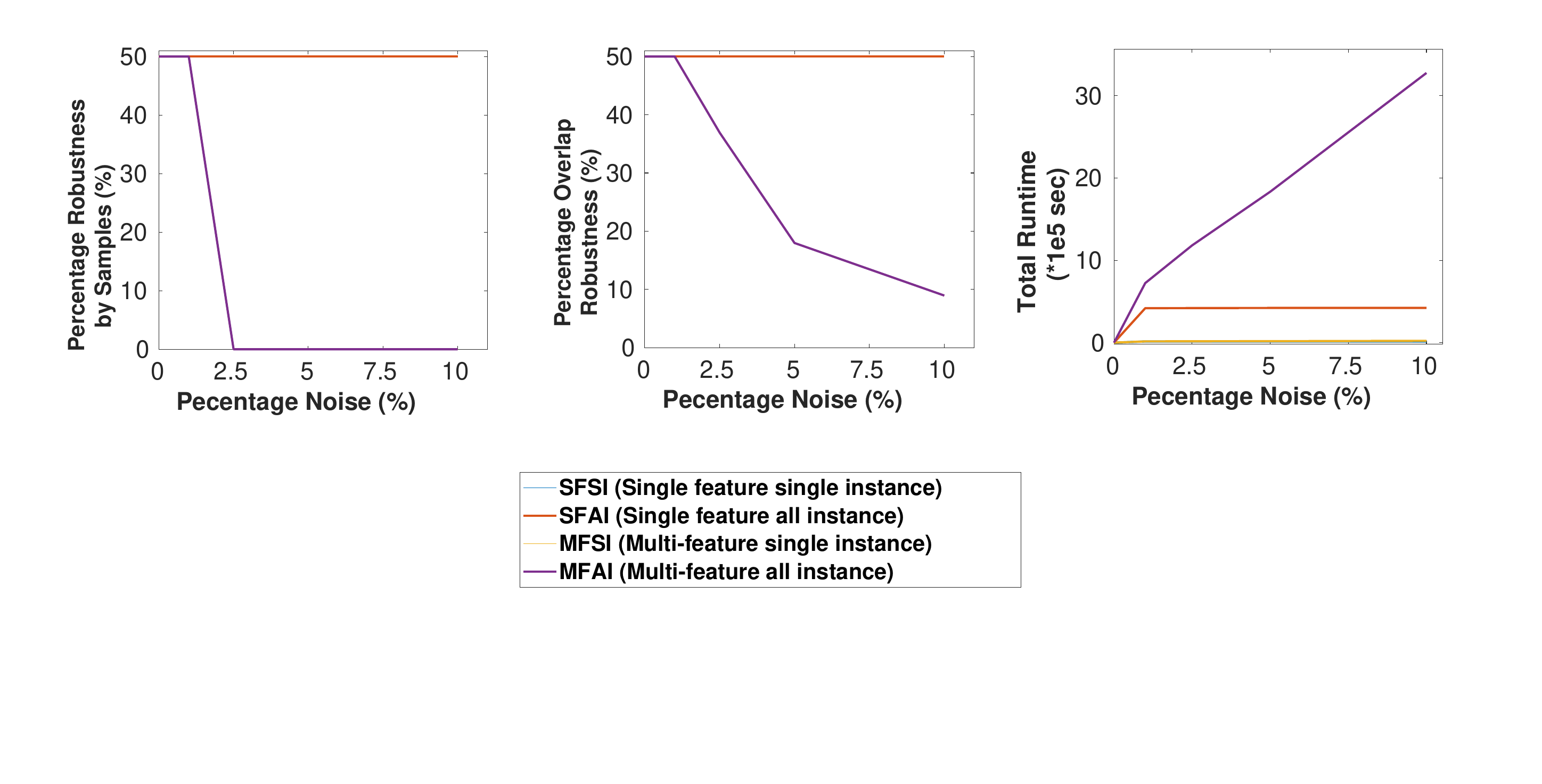}
    \vspace*{-\baselineskip}
    \vspace*{-\baselineskip}
    \vspace*{-\baselineskip}
    \caption{Percentage Robustness and Runtime plots w.r.t increasing noise}
    \label{fig:SOCRobustnessplot}
    \vspace*{-\baselineskip}
    \vspace*{-\baselineskip}
\end{figure}
\vspace*{-\baselineskip}
\subsubsection{Observation and Analysis:}
Fig.~\ref{fig:BSOCBounds} shows a sample plot for gradually increasing estimation bounds with increasing MFSI noise. We can see from the figure that for each time instance, the system becomes locally non-robust as the noise value increases.

Table.~\ref{tab:Table2} presents the network's overall performance, i.e., \newblue{the percentage robustness measures, PR [Def.~\ref{def:PSR}], POR [Def.~\ref{def:POR}] and average verification runtime (avgRT)}, with respect to each noise. The percentage robustness values start decreasing and the average (as well as total) runtime starts increasing as the measure of noise increases for MFAI and MFSI, but for SFSI and SFAI it remains the same for these noise perturbations considered. This is because in the first case, the noise is added to all the features, resulting in increasing the cumulative effect of disturbance on the output estimation. However, in the other case, the noise is attached only to a single feature, assuming that not all features will get polluted by noise simultaneously; and that the reachable bounds are in the acceptable range. A plot of robustness values and the total runtime is shown in Fig~\ref{fig:SOCRobustnessplot}.

We can also see that the decrease in POR values for MFSI and MFAI are less compared to the PR values with increasing noise because, for PR calculation, only those time steps are considered where the estimated range falls entirely within the allowed range, whereas for POR calculation even if some part of the estimated range goes outside the allowable range, their fractional contribution is still considered.

Another interesting observation here is the robustness matrices for both SFSI and SFAI are the same; however, the computations for SFAI take almost three times longer than the computations for SFSI. The same analogy is observed for MFSI and MFAI datasets but with an even higher time taken for MFAI. The possible reason for this observation could be that, while the data is subjected to perturbations across all time instances, the noise added to the final time step has the most significant impact on the output.

\paragraph{ \textbf{Turbofan Engine Degradation Simulation Data Set (TEDS)}:}
 In this dataset, the acceptable RUL bounds are considered to be $\pm 10$ of the actual RUL. For instances where the lower bound is negative, we assume those values to be 0 as well. We then calculate the percentage robustness measures, PR [Def.\ref{def:PSR}], POR [Def.\ref{def:POR}], and \newblue{average verification runtime (avgRT)}, for an input set with all 100 consecutive data points, each having 30 time instances. The results for three different noise values, $0.1\%$, $0.5\%$, and $1\%$ of the mean ($\mu$), are presented in Table~\ref{tab:Table4}. For SFSI and SFAI noises, we randomly choose a feature (feature 7, representing sensor 2) for noise addition. The noise is added to the last time step ($t_{30}$) of each data sample for SFSI and SFAI noises. The results of the MFAI noise have been omitted due to scalability issues, as it is computationally heavy and time-consuming \footnote{\newblue{The MFAI noise, i.e., adding the $L_\infty$ norm to all feature values across all time instances, significantly increases the input-set size compared to other noise types. This leads to computationally expensive calculations for layer-wise reachability, resulting in longer run times. Moreover, noise in an industrial setting affecting all features over an extended period is unlikely. Considering these factors, we decided to exclude the results of the MFAI noise for the TEDS dataset from our analysis.}}.
 
\begin{figure}[h!]
    \centering
    \includegraphics[width = 0.75\columnwidth]{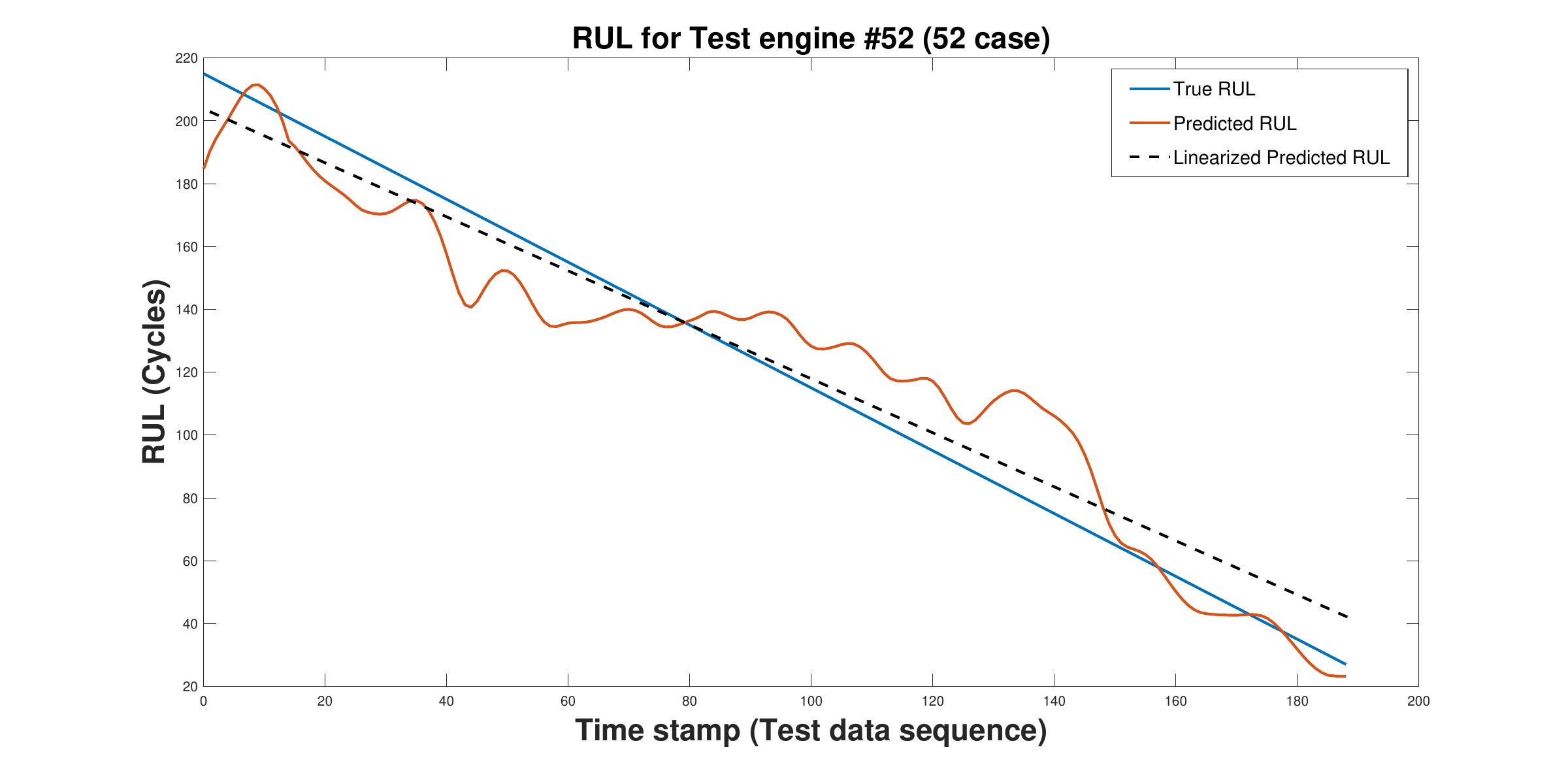}
    \vspace*{-\baselineskip}
    % \vspace*{-\baselineskip}
    % \vspace*{-\baselineskip}
    \caption{Percentage Robustness and Runtime plots w.r.t increasing noise}
    \label{fig:localmonotonicity}
    \vspace*{-\baselineskip}
    \vspace*{-\baselineskip}
\end{figure}

For verifying the local monotonicity of the estimated output RUL bounds at a particular time instance, we have fitted the previous RUL bounds along with the estimated one in a linear equation as shown in Fig. \ref{fig:localmonotonicity}. This guarantees the monotonically decreasing nature of the estimated RUL at any time-instance.
\vspace*{-\baselineskip}
\vspace*{-\baselineskip}
\begin{table}[h!]
    \caption{Global Robustness: Percentage Robustness(PR) for noises for 100 consecutive time steps}
    \label{tab:Table4}
    \centering 
    \begin{tabular}{ccccccc}
    \toprule
    \centering
    $noise$ & $PR_{SFSI}$ & $POR_{SFSI}$ & $avgRT_{SFSI}(s)$ & $PR_{SFAI}$ & $POR_{SFAI}$ & $avgRT_{SFAI}(s)$\\ 
    \hline 
    1 & 13 & 13 & 1.0796 & 13 & 13.31 & 32.8670\\ 
    2.5 & 13 & 13 & 1.1755 & 12 & 13.13 & 62.1483\\
    5 & 13 & 13 & 1.2908 & 8 & 12.64 & 108.0736\\
    % 10 & 100 & 100 & 0.7027 & 100 & 100 & 21.0780\\
    \hline 
    \end{tabular}
    \begin{tabular}{ccccccc}
    \toprule
    \centering
    $noise$ & $PR_{MFSI}$ & $POR_{MFSI}$ & $avgRT_{MFSI}(s)$ \\%& $PR_{MFAI}$ & $POR_{MFAI}$ & $avgRT_{MFAI}(s)$\\ 
    \hline 
    1 & 13 & 13 & 9.6567 \\%& 100 & 100 & 36.1723 
    2.5 & 13  & 13 & 10.2540 \\%& 0  & 73.87 & 59.0588
    5 & 13 & 13 & 11.2100 \\%& 0 & 35.95 & 91.6481
    % 10 & 0 & 17.89 & 1.1051 & 0 & 17.89 & 163.7568\\
    \hline 
    \end{tabular}
    \vspace*{-\baselineskip}
    \vspace*{-\baselineskip}
\end{table}
\vspace*{-\baselineskip}
% \vspace*{-\baselineskip}
\begin{figure}[h!]
    \centering
    \includegraphics[width = 0.9\columnwidth]{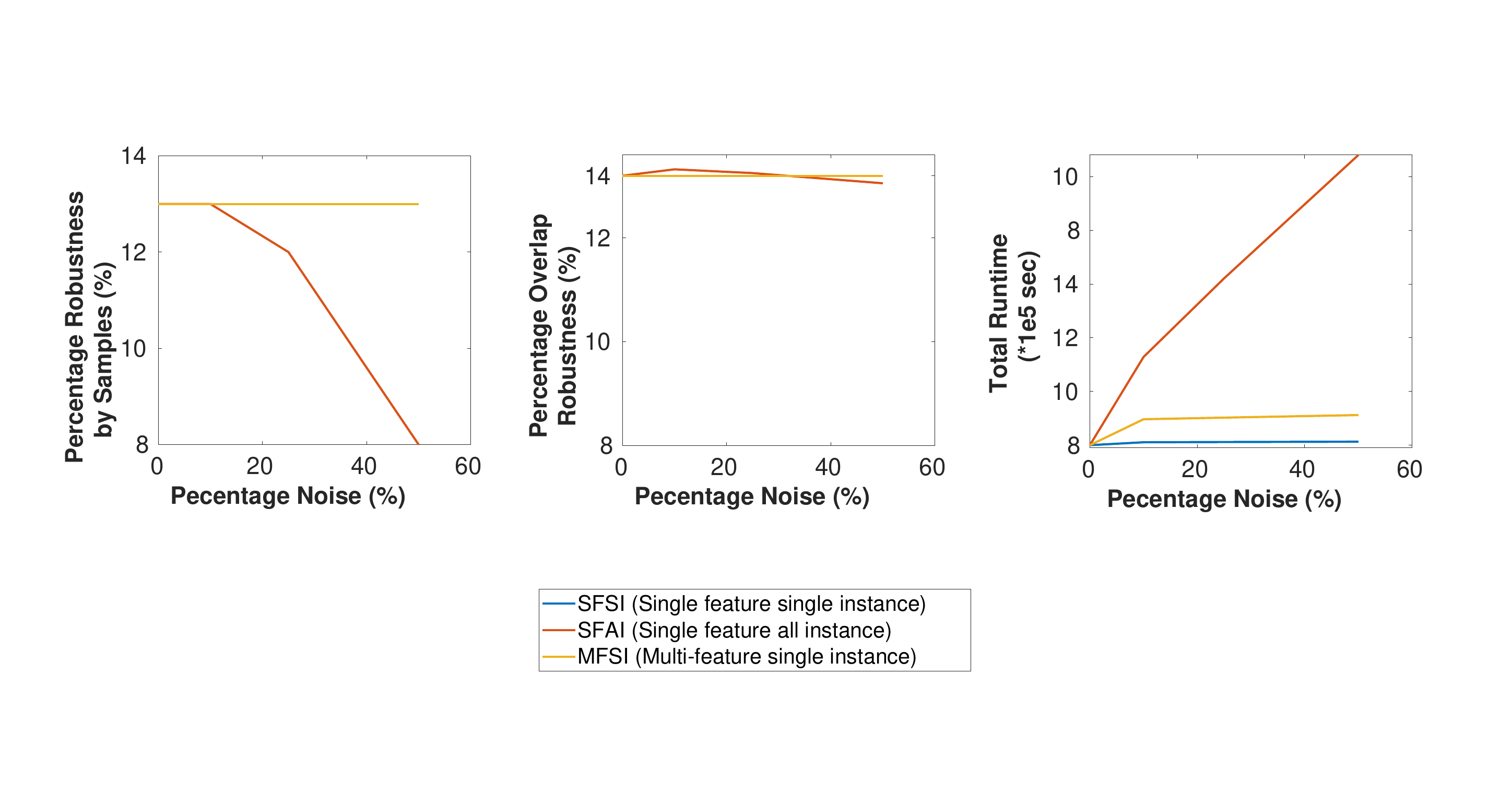}
    \vspace*{-\baselineskip}
    \vspace*{-\baselineskip}
    \vspace*{-\baselineskip}
    \caption{Percentage Robustness and Runtime plots w.r.t increasing noise}
    \label{fig:RULRobustnessplot}
    \vspace*{-\baselineskip}
\end{figure}
\vspace*{-\baselineskip}
\subsubsection{Observation and Analysis:}Fig.~\ref{fig:RULBounds} shows a sample plot for gradually increasing estimation bounds with increasing SFAI noise. Here we need to notice that the network's performance in terms of following the actual RUL value is not well. However, Table.~\ref{tab:Table4} presents the network's overall performance with respect to each noise. Contrary to the other dataset, we see that the percentage robustness measures corresponding to SFAI and SFSI noises differ. Interestingly, while the noise value increases, the PR, and POR for SFSI remain the same, whereas the robustness measures for SFAI decrease. However, the performance matrices for MFSI are the same as the SFSI except for the time. This might be because, for both SFSI and MFSI, the noise is added only at a single time instance, whereas for SFAI, the noise is added to the entire time instances, resulting in an increased cumulative effect of disturbance on the output. 

\newblue{Our results consistently show higher POR values than PR values in Table. [\ref{tab:Table2}-\ref{tab:Table4}]. Since we assess output reachable bounds using $L_\infty$ perturbations in the input, we acknowledge the significance of cases where reachable sets overlap with permissible bounds but do not entirely fall within them. In summary, PR measures adopt a more conservative approach, while POR captures the relationship between output reachable bounds and permissible bounds more accurately.}

%% file: 7_conclusion.tex
\section{Conclusion and Future Work}
\label{sec: Conclusion and Future Work}
This paper explores formal method-based reachability analysis of variable-length time series regression neural networks (NNs) using approximate Star methods in the context of predictive maintenance, which is crucial with the rise of Industry 4.0 and the Internet of Things. The analysis considers sensor noise introduced in the data. Evaluation is conducted on two datasets, employing a unified reachability analysis that handles varying features and variable time sequence lengths while analyzing the output with acceptable upper and lower bounds. Robustness and monotonicity properties are verified for the TEDS dataset. Real-world datasets are used, but further research is needed to establish stronger connections between practical industrial problems and performance metrics. The study opens new avenues for exploring perturbation contributions to the output and extending reachability analysis to 3-dimensional time series data like videos. Future work involves verifying global monotonicity properties as well, and including more predictive maintenance and anomaly detection applications as case studies. \newblue{The study focuses solely on offline data analysis and lacks considerations for real-time stream processing and memory constraints, which present fascinating avenues for future research.}
\paragraph{\textbf{Acknowledgements.}}
The material presented in this paper is based upon work supported by the National Science Foundation (NSF) through grant numbers 1910017, 2028001, 2220418, 2220426, and 2220401, and the Defense Advanced Research Projects Agency (DARPA) under contract number FA8750-18-C-0089 and FA8750-23-C-0518, and the Air Force Office of Scientific Research (AFOSR) under contract number FA9550-22-1-0019 and FA9550-23-1-0135. Any opinions, findings, conclusions, or recommendations expressed in this paper are those of the authors and do not necessarily reflect the views of AFOSR, DARPA, or NSF. We also want to thank our colleagues, Tianshu and Barnie for their valuable feedback.

%% file: 10_appendix.tex
\newpage
\appendix
\section{Appendix} \label{Appendix}
% \subsection{Sigma Levels (\textbf{Figure} \ref{fig:Sixsigma})}
% \begin{figure}[h!]
%     \centering
%     \includegraphics[width = 0.99 \columnwidth]{Images/10_appendix/six_sigma_normal_distribution_2.jpeg}
%     \caption{Six-sigma distribution and 3-sigma rule [source: Google].}
%     \label{fig:Sixsigma}
% \end{figure}
% \vspace*{-\baselineskip}
% \vspace*{-\baselineskip}
\subsection{Preliminaries}
\begin{definition}[Soundness]\label{def: soundness}
Let $\mathcal{F}: R^j \rightarrow R^p$ be a NN with an input set $R_0$ and output reachable set $R_f$ . The
computed $R_f$ given $\mathcal{F}$ and $R_0$ is sound iff $\forall{x} \in R_0, | y = \mathcal{F}(x), y \in R_f$.
\end{definition}
\begin{definition}[Completeness]\label{def: completeness}
    Let $\mathcal{F}: R^j \rightarrow R^p$ be a NN with an input set $R_0$ and output reachable set $R_f$ . The
computed $R_f$ given $\mathcal{F}$ and $R_0$ is complete iff $\forall{x} \in R_0, \exists{y} = \mathcal{F}(x)~|~y \in R_f$ and $\forall{y} \in R_f, \exists{x} \in R_0~|~y = \mathcal{F}(x)$.
\end{definition}
\subsubsection{$L_\infty$ Norm:}\label{l_inf norm} Input perturbations can be quantified using different types of norms. Here, we have used the $L_\infty$ norm, which records the greatest perturbation magnitude among all input elements.
\begin{equation}
    L_\infty : ||x - x'||_\infty = max||x_i - x'_i||
\end{equation}
\subsubsection{Example of Robustness Calculations:}\label{Robustness Measures Example}
The left picture of Fig. \ref{fig:LocalRobust} in Appendix depicts an example plot of output estimations (red) vs. the allowable bounds (blue). Here, we can see that the network is locally robust for time instances $t_1$ and $t_5$; in other instances, it is non-robust w.r.t the noise added. So the RV is 1 for both $t_1$ and $t_5$ and 0 for others. 
\vspace*{-\baselineskip}
\begin{figure}[h!]
    \centering
    \includegraphics[width = 0.8\columnwidth]{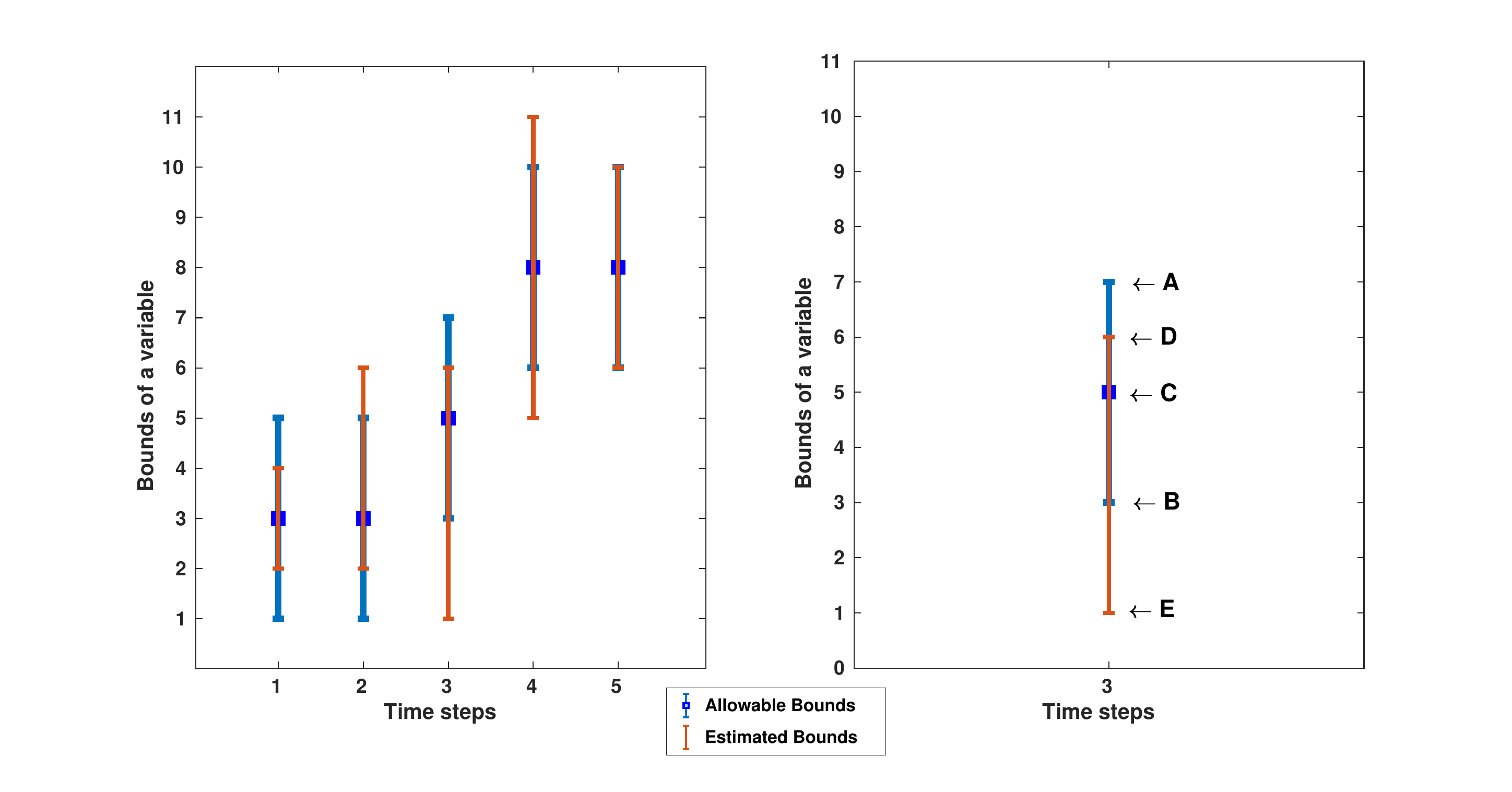}
    \caption{Example: (left)Output estimation bounds (red) of a TSRegNN and Allowable (blue) over five consecutive time step
    and (right) for a particular time step $t_3$}
    \label{fig:LocalRobust}   
    \vspace*{-\baselineskip}
\end{figure}
To better understand the concept of POR and PO, we refer to the right picture of Fig. \ref{fig:LocalRobust}. Here for a time instance $t_3$, C denotes the actual signal value, AB is the allowable output range, and DE is the estimated reachable bounds. Here, $N_{robust}$ = 3 and $N_{total}$ = 5, so the PR for this particular example is 60\%.
The PO will be calculated as:
\begin{equation*}
 PO = \frac{DB}{DE} = \frac{3}{5} = 0.6
\end{equation*}
To calculate POR, we calculate the PO for each of the time instances:
\begin{equation*}
 POR = \frac{\frac{2}{2}+\frac{3}{4} + \frac{3}{5} + \frac{4}{6} + \frac{4}{4}}{5} \times 100\% = 80.33\% (approx)
\end{equation*}

\subsection{Battery State-of-charge}\label{Sec: State-of-charge} Battery state-of-charge is a measurement of the amount of energy available in a battery at a specific point in time, expressed as a percentage. This term is often used in various applications involving battery-powered systems, e.g., electric vehicles, renewable energy storage systems, portable electronics etc. Accurate estimation of the State of Charge (SOC) of a battery is crucial for efficient battery management and ensuring the longevity of the battery. The SOC is expressed as a percentage of the full capacity of the battery.

\subsection{Remaining Useful Life}\label{Sec: Remaining useful Life} The Remaining Useful Life (RUL) is a subjective estimate of the lifespan of any equipment before it requires repair or replacement. This important concept is often used in various fields, including maintenance, reliability engineering, and prognostics and health management (PHM). RUL estimation is typically based on the analysis of historical data, such as sensor measurements, degradation patterns, maintenance records, and operational conditions. Various techniques and models, including statistical methods, machine learning algorithms, and physics-based approaches, are generally used to predict the RUL.

\subsection{Sample Dataset Plots}\label{AppendixSampleDataset}
\begin{figure}
    \centering
    \includegraphics[width = 0.8\columnwidth]{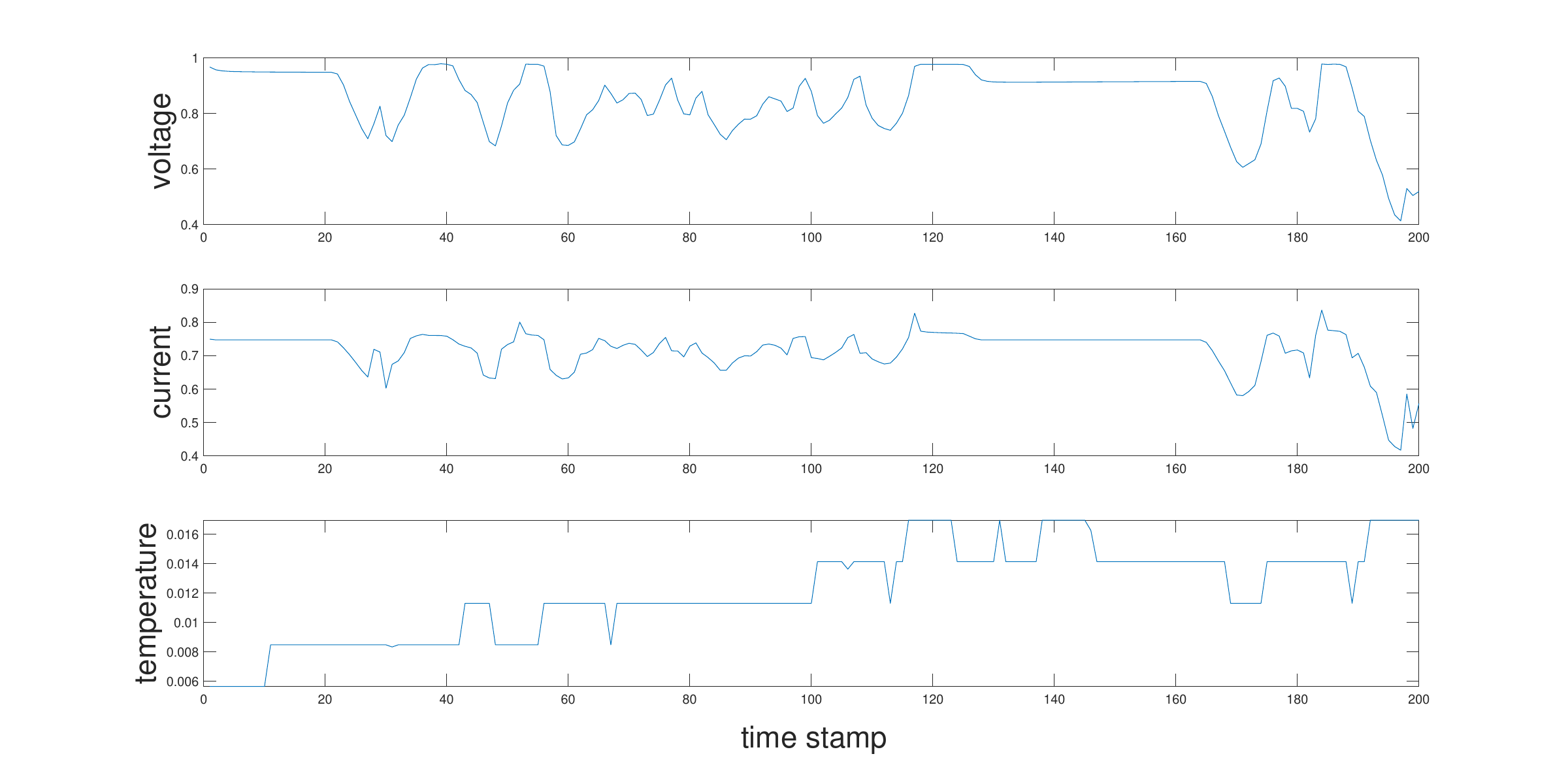}
    \caption{Sample feature value plot for Battery SOC Dataset.}
    \label{fig:SampleSFDataset}
\end{figure}
\begin{figure}[h!]
    \centering
    \includegraphics[width = 0.7\columnwidth]{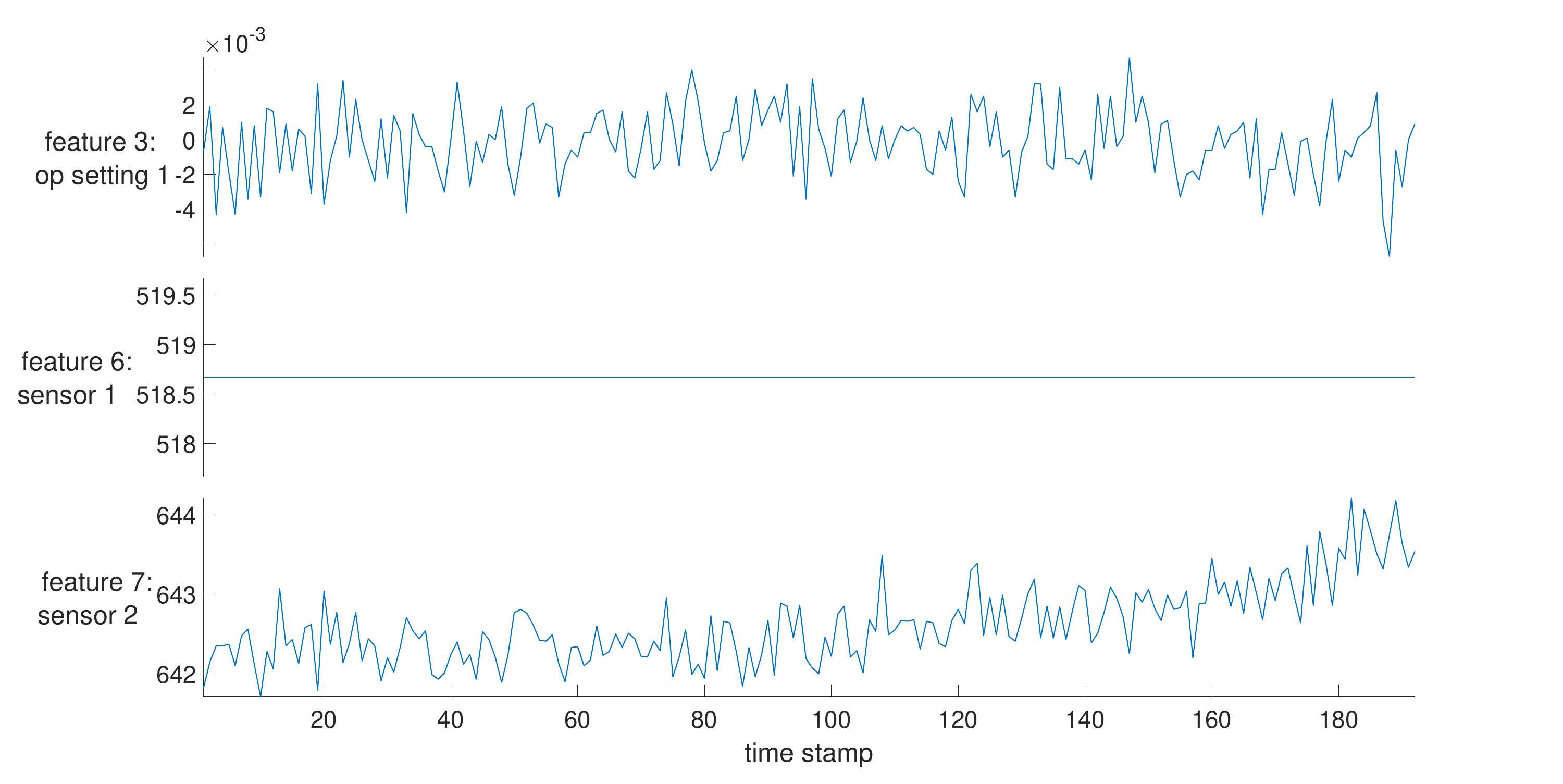}
    \caption{Sample feature value plot for Turbofan Engine Degradation Simulation Dataset.}
    \label{fig:SampleTEDSataset}
\end{figure}
\subsection{Prognosability}\label{AppendixPrognosability} Features that remain constant for all time steps can negatively impact the training. Feature reduction is done using the 'prognosability' MATLAB command. Prognosability is actually a property relative to the prediction of the future state of the system, and this term is mainly used for lifetime data. In MATLAB, 'prognosability' is used as a function to measure the variability of the features in a dataset at failure. The equation for the prognosability calculation is given as below:
\begin{equation}
    prognosability = Y = \exp{\frac{std_j(x_j(N_j))}{mean_j\lvert(x_j(1) - x_j(N_j)\rvert}}
\end{equation}
The output has 3 different outcomes:
\begin{enumerate}
    \item Y = 0 means the feature values are constant, i.e., no variability in the data.
    \item Y = NaN indicates the prognosability could not be calculated.
    \item Y = 1 means the feature values are perfectly prognosable i.e., there is variability in the data.
\end{enumerate}
\subsection{Feature Reduction of TEDS Dataset}\label{AppendixTEDSDataset} 
Each engine has 26 different feature
values, recorded at different time instances.
\begin{enumerate}
        \item Feature 1: Unit number
        \item Feature 2: Time-stamp
        \item Feature 3–5: Operational settings
        \item Feature 6–26: Sensor measurements 1–21
    \end{enumerate}
After analyzing the dataset using 'prognosability', number of features considered for NN training reduced to 17 from 26, and they are
\begin{enumerate}
        \item Feature 3–4~~~~: Operational settings 1-2
        \item Feature 7–9~~~~: Sensor measurements 2-4
        \item Feature 11–14~: Sensor measurements 6–9
        \item Feature 16–20~: Sensor measurements 11–15
        \item Feature 22~~~~~~: Sensor measurements 17
        \item Feature 25-26~~: Sensor measurements 20-21
\end{enumerate}

\begin{figure}[htb!]
    \centering
    \includegraphics[width = \columnwidth]{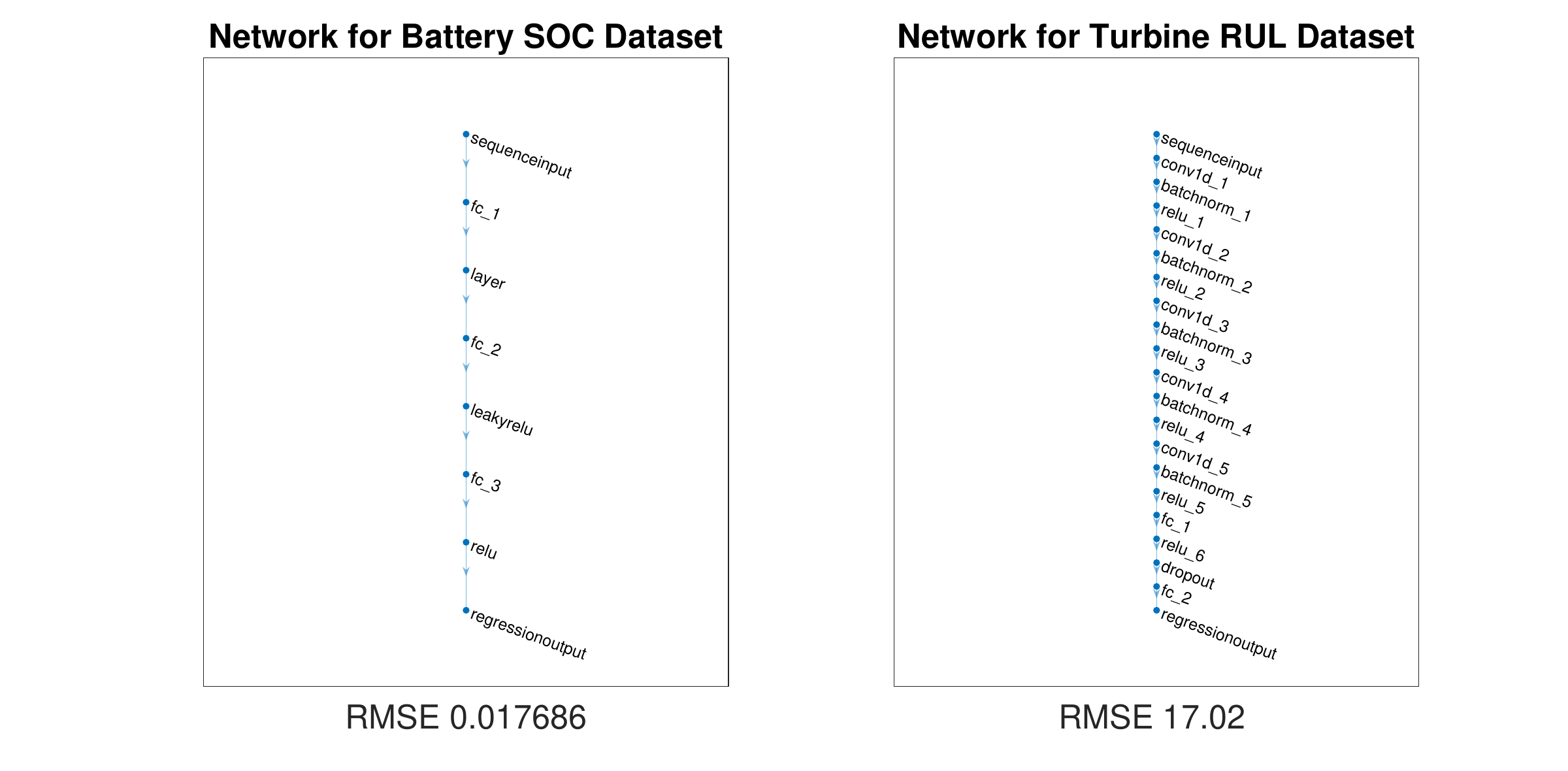}
    \caption{The architectures of the regression networks for both the datasets.}
    \label{fig:FigArchitecture}
\end{figure}

\newpage
\subsection{Network Architectures: (\textbf{Figure}\label{Appendix A.1} \ref{fig:FigArchitecture})}\label{AppendixNetworks}
\subsection{Noise}\label{AppendixNoise}
\begin{figure}[htb!]
    \centering
    \includegraphics[width = 0.8\columnwidth]{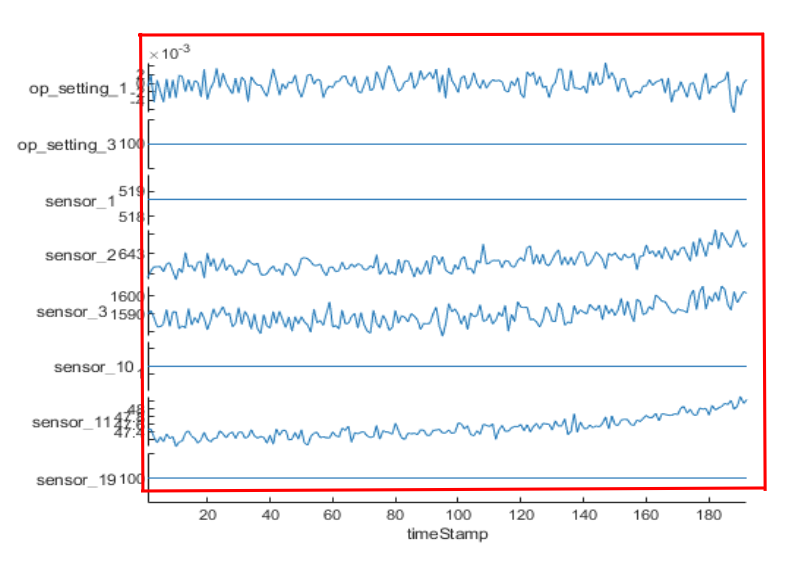}
    \caption{MFAI Noise: noise added to the complete time series data, for all features.}
    \label{fig:MFAI}
\end{figure}
\begin{figure}[htb!]
    \centering
    \includegraphics[width = 0.9\columnwidth]{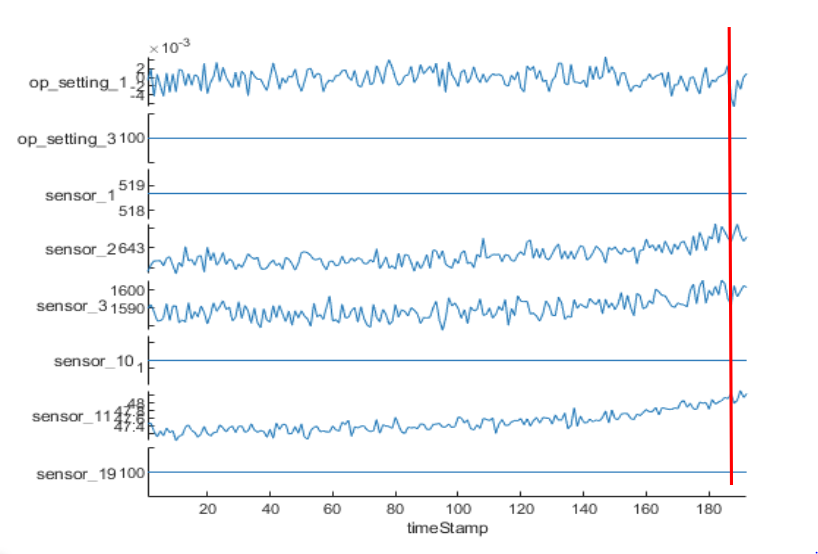}
    \caption{MFSI Noise: noise added at a particular instance of the time series data, for all features.}
    \label{fig:MFSI}
\end{figure}
\begin{figure}[htb!]
    \centering
    \includegraphics[width = 0.9\columnwidth]{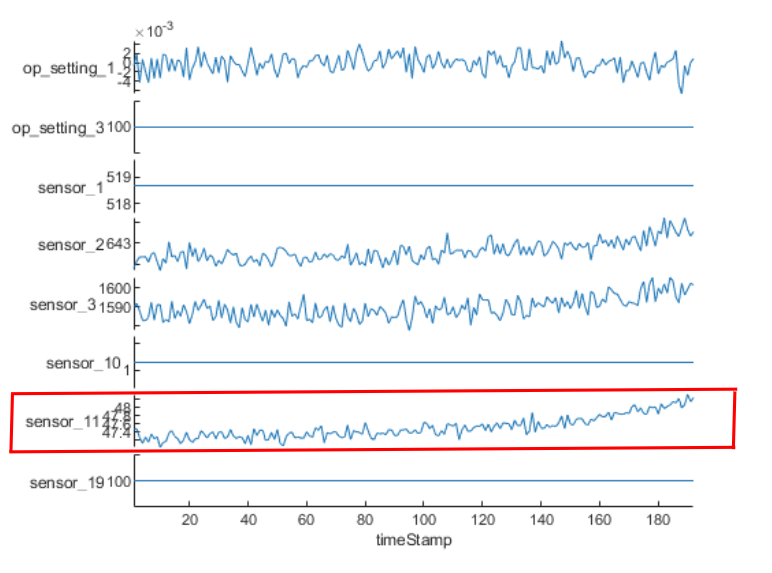}
    \caption{SFAI Noise: noise added to the complete time series data, a particular feature.}
    \label{fig:SFAI}
\end{figure}
\begin{figure}[htb!]
    \centering
    \includegraphics[width = 0.85\columnwidth]{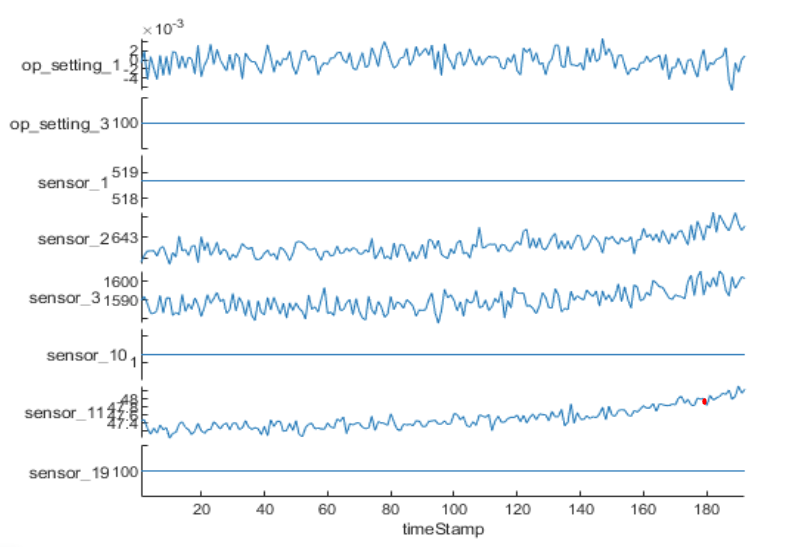}
    \caption{SFSI Noise: noise added at a particular instance of the time series data, a particular feature.}
    \label{fig:SFSI}
\end{figure}

\subsection{TEDS dataset Sample Reachability Plot}
\begin{figure}[h!]
    \centering
    \includegraphics[width = 0.9\columnwidth]{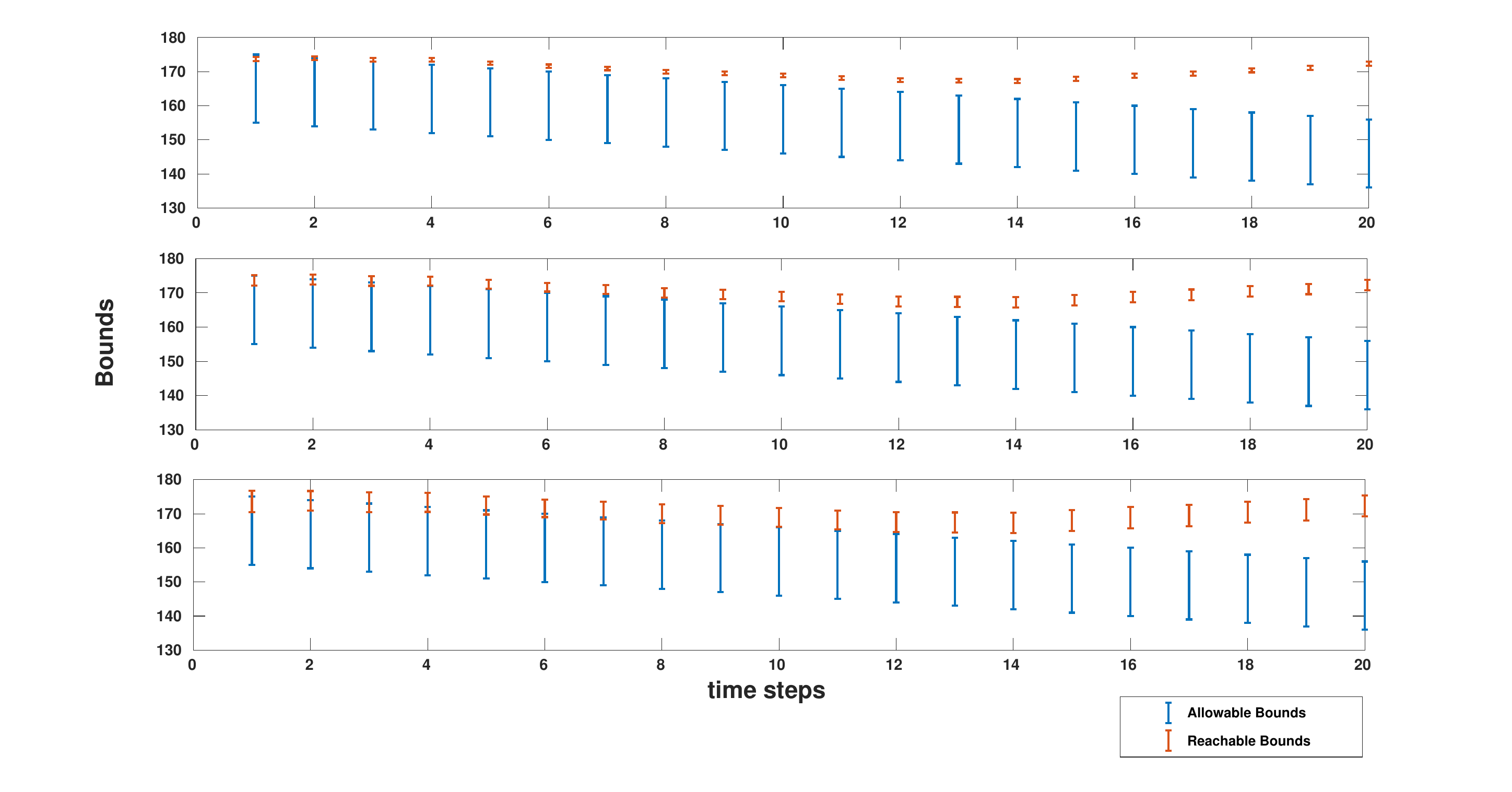}
    \vspace*{-\baselineskip}
    % \vspace*{-\baselineskip}
        \caption{Allowable (blue) and reachable (red) bounds for battery SOC dataset for 100 consecutive time steps and three different SFAI noise values $1\%$ (upper), $2.5\%$ (middle) and $5\%$ (lower) respectively }
    \label{fig:RULBounds}
    \vspace*{-\baselineskip}
\end{figure}